\documentclass[11pt]{article}
\usepackage[utf8]{inputenc}
\usepackage[T1]{fontenc}
\usepackage{lmodern}
\usepackage[margin=1in]{geometry}
\usepackage{amsmath,amssymb}
\usepackage{graphicx}
\usepackage{float}
\usepackage{booktabs}
\usepackage{longtable}
\usepackage{array}
\usepackage{caption}
\usepackage{marvosym}
\usepackage[authoryear,round]{natbib}
\usepackage[hidelinks]{hyperref}
\usepackage{url}

\newcommand{\Sbar}{\bar{S}}
\newcommand{\ebar}{\bar{e}}
\newcommand{\Rpt}{\mathrm{Rpt}}
\newcommand{\Rpd}{\mathrm{Rpd}}

\begin{document}

\begin{center}
{\Large\bfseries Semantic repeatability and reproducibility of GPT-5 and Gemini 3 Flash on H5N1 One Health questions\par}
\vspace{1em}
\textbf{Luizza Dalla Pria}\textsuperscript{1,2}~\Letter, \textbf{Amelia L. Nelson}\textsuperscript{2} \& \textbf{Renata Ivanek}\textsuperscript{2}
\end{center}

\noindent\textsuperscript{1}College of Professional Studies, Northeastern University, Boston, MA, USA.

\noindent\textsuperscript{2}Department of Population Medicine and Diagnostic Sciences, College of Veterinary Medicine, Cornell University, Ithaca, NY, USA.


\section*{Abstract}

Highly pathogenic avian influenza A(H5N1) has re-emerged as a One Health concern since 2024, and the public is increasingly consulting large-language-model (LLM) chatbots for health information during outbreaks. Before a chatbot's accuracy can be meaningfully assessed, its consistency must first be understood, so that any single response can be trusted as a fair representative of the model's behavior. We characterized the semantic repeatability and reproducibility of GPT-5 (OpenAI) and Gemini 3 Flash (Google) on 15 H5N1 One Health questions sourced from social media discourse by adopting and extending a centroid-based evaluation framework. Each question was submitted 30 times and reformulated using five Chain-of-Thought (CoT) variants with 30 runs each, yielding 5,400 responses. Pairwise cosine similarity was computed on MPNet embeddings, with sensitivity analyses across three additional transformers. Statistical inference combined standard nonparametric tests (paired Wilcoxon and Friedman with Holm--Bonferroni correction) with inter-embedding agreement statistics (ICC(2,k) and Cronbach's $\alpha$) introduced here as a methodological addition. Both models produced high mean consistency (0.93--0.97, rescaled to {[}0, 1{]}). GPT-5 showed reproducibility significantly higher than repeatability (W = 3.0, p = 0.0003), whereas for Gemini 3 Flash the two metrics were statistically indistinguishable (W = 42.0, p = 0.33). Differential Diagnosis ranked among the top two CoT framings for consistency in both models. The tested chatbots behave as semantically stable sources on H5N1 One Health questions, and the developed consistency evaluation framework is portable to other domains.

\section*{Author Summary}

During a disease outbreak, people increasingly turn to AI chatbots such as ChatGPT or Gemini for answers. For these tools to be useful as a public information source, they should at least give consistent answers when asked the same health question more than once, or when the question is reworded slightly. If the answers vary, the chatbot cannot be reliably evaluated for accuracy, and the public may receive different information depending on how they phrase their query. We tested two widely used chatbots from leading providers, GPT-5 (OpenAI) and Gemini 3 Flash (Google), on 15 real questions about H5N1 bird flu drawn from public online discussions. Each question was asked 30 times in its original form and in five rephrased forms based on common reasoning styles. We measured how similar the chatbot's answers were to each other using sentence embedding models. Both chatbots produced highly consistent answers about H5N1 overall, though the degree of consistency depended on how the question was phrased and differed between the two models. Our approach can be reused for other health topics or different domains, and establishes the groundwork needed before the accuracy of a chatbot can be meaningfully assessed by expert reviewers.

\section*{Introduction}

The ongoing highly pathogenic avian influenza A(H5N1) outbreak in the United States has produced sustained public interest in health information since 2024, with confirmed human cases continuing to accumulate (CDC, 2026) and continued outbreaks reported in dairy cattle and poultry across multiple states (USDA APHIS, 2026). In parallel, members of the public increasingly turn to large language model (LLM)-based chatbots (e.g., ChatGPT, Gemini) as a first stop for health-related information (Bautista et al., 2025), and the quality of those chatbot responses has itself become a focus of empirical comparison with physician answers (Ayers et al., 2023). When the public uses these chatbots to ask about emerging zoonotic diseases, whether the chatbots give consistent answers becomes a public-health-relevant question --- and a necessary prerequisite to any later assessment of correctness, since an LLM that gives different outputs each time cannot be meaningfully evaluated for accuracy on a single output. This study assesses the consistency of two leading commercial LLM-based chatbots --- OpenAI's GPT-5 and Google's Gemini 3 Flash --- when answering H5N1-related One Health questions sourced from real public discourse. These two models were selected as widely used offerings from two different major providers, with closely spaced release dates that support direct cross-provider comparison: GPT-5 on August 7, 2025, and Gemini 3 Flash on December 17, 2025, both made available during the ongoing H5N1 outbreak that began in 2024.

Measuring whether a chatbot gives consistent outputs to the same prompt or to variants of the same prompt requires a working definition of ``semantic consistency.'' Two responses can use different words while conveying the same meaning (semantically consistent, but lexically dissimilar), or two that are worded similarly can differ substantively in meaning (semantically inconsistent, but lexically similar). In this study, we therefore take a meaning-level view semantic consistency, under which two responses count as consistent when they communicate the same underlying content, regardless of their exact wording. We measure two complementary dimensions of semantic consistency. The first, repeatability, asks whether the chatbot gives similar semantic answers when it is asked the exact same question multiple times. The second, reproducibility, asks whether the chatbot gives similar semantic answers when the underlying question is rephrased through different Chain-of-Thought (CoT) reasoning framings --- prompting styles that explicitly instruct the model to work through a question using a particular reasoning approach (e.g., step-by-step reasoning, differential diagnosis, or Bayesian updating) before producing its answer (Savage et al., 2024). CoT framings matter because the way a question is framed can substantially influence the content of LLM outputs; measuring reproducibility across different CoT framings therefore tests how consistent the chatbot's semantic content is to the user's phrasing choices. Both dimension are formalized in the statistical framework of Shyr et al. (2025), which we adopt and extend here.

Prior empirical work on LLM semantic consistency has concentrated in medical and clinical domains. Shyr et al. (2025) demonstrated their framework on clinical diagnostic reasoning tasks using ChatGPT-4, ChatGPT-4o-mini, and Llama 3.2-1B. Hackl et al. (2023) documented high consistency for GPT-4 on a non-clinical task: student response grading. Li et al. (2024) showed that prompt-engineering style alone can span the full range of inter-rater agreements on orthopedic clinical guidelines evaluated with GPT-4-Web and GPT-3.5. Collectively, these studies establish that LLM consistency is both measurable and sensitive to prompt and model choice. We extend this work along four complementary axes: (i) content domain, (ii) embedding-model breadth, (iii) cross-provider scope, and (iv) statistical reporting. For (i) content domain, we moved from clinical and specialty-medical material to One Health questions sourced from public discourse. For (ii) embedding-model breadth, we substituted a general-purpose sentence-embedding model for Shyr et al.'s single clinical-domain embedding, and measured consistency across multiple sentence-embedding models as a sensitivity check. For (iii) cross-provider scope, we benchmarked two commercial LLMs from different companies (OpenAI's GPT-5 and Google's Gemini 3 Flash) rather than variants from a single provider. For (iv) statistical reporting, we extended Shyr et al.'s nonparametric comparison pipeline with paired tests for the within-model repeatability-vs-reproducibility comparison, sampling uncertainty of per-prompt means, and inter-embedding agreement statistics. The H5N1 One Health questions offer a useful test case for addressing these gaps: the topic draws active public interest during the ongoing outbreak introduced above, the One Health framing spans human, animal, and environmental health, and our questions were sourced from real online public discourse rather than researcher-constructed vignettes.

Therefore, the present study was designed around three objectives. First, we quantify the semantic repeatability and reproducibility of two contemporary commercial LLMs --- GPT-5 (OpenAI) and Gemini 3 Flash (Google) --- when answering naturalistic H5N1 One Health questions sourced from publicly available online discussions. Second, we assess whether patterns of consistency differ systematically across the two commercial LLM providers and across five CoT reasoning framings. Third, we evaluate whether the centroid-based semantic consistency framework of Shyr et al. (2025), originally developed on clinical diagnostic reasoning tasks, transfers to a public-health content domain and across a broader set of sentence-embedding models.

\section*{2. Methods}

\subsection*{2.1 Study design and context}

This section describes the design, data, and analyses underlying the study. Table 1 defines the key semantic-consistency concepts used throughout the manuscript, and Figure 1 provides a visual overview of the six-step analysis pipeline.

\begin{figure}[H]
\centering
\includegraphics[width=\textwidth]{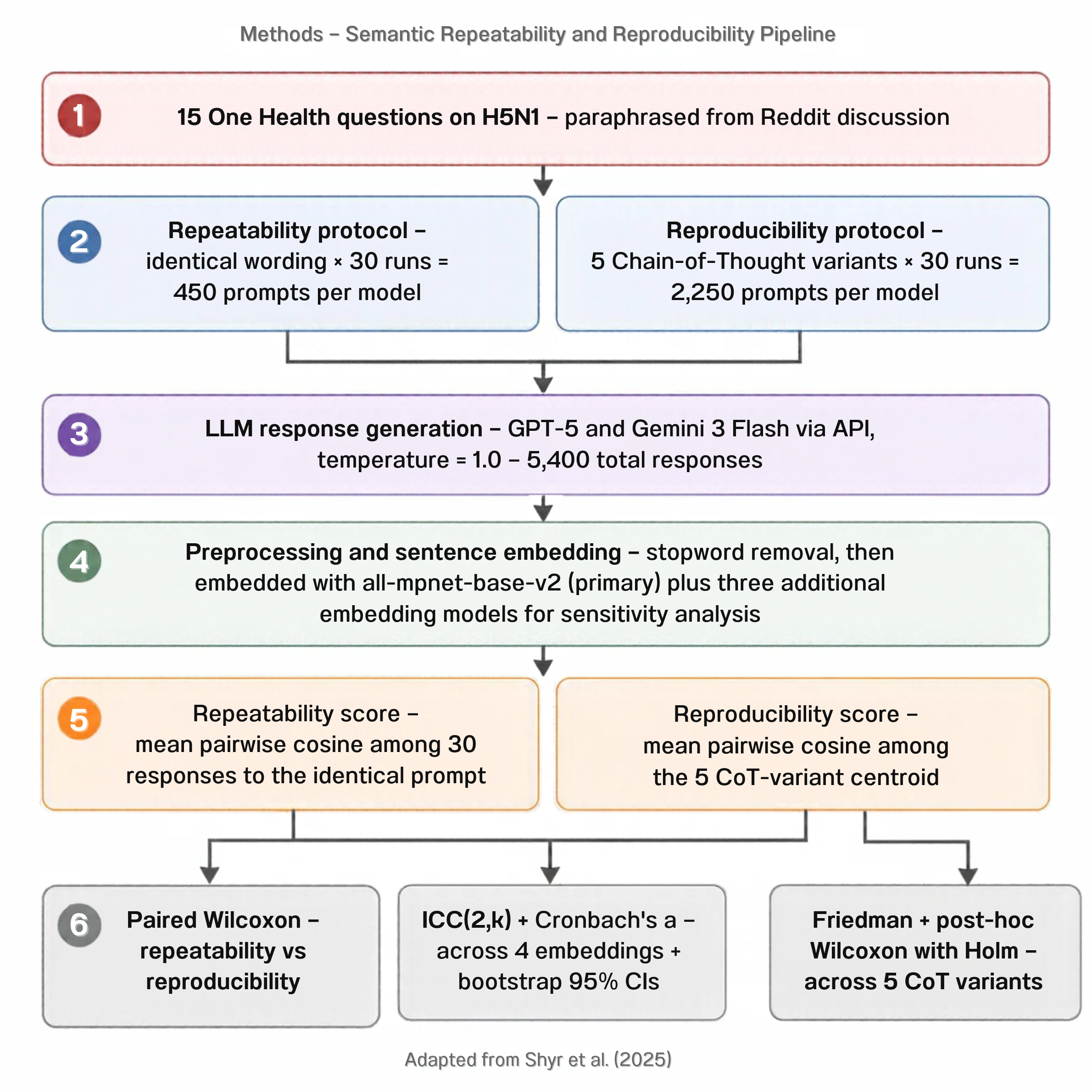}
\caption{\textbf{Figure 1. Methods pipeline.}}
\label{fig:pipeline}
\end{figure}

\refstepcounter{table}\label{tab:concepts}%
\noindent \textbf{Table 1.} Key semantic-consistency and methodological concepts used in this study.

{\small
\begin{longtable}[]{@{}>{\raggedright\arraybackslash}p{0.20\linewidth}>{\raggedright\arraybackslash}p{0.54\linewidth}>{\raggedright\arraybackslash}p{0.20\linewidth}@{}}
\toprule
\endhead
\textbf{Concept} & \textbf{Definition} & \textbf{Reference}\tabularnewline
\textbf{Prompt} & The text input given to a large language model (LLM) to elicit a response. In this paper, ``\emph{prompt}'' and ``\emph{question}'' are used interchangeably --- both refer to the text sent to the LLM. & Liu et al. (2023)\tabularnewline
\textbf{Model} & Throughout this paper, ``model'' refers to a LLM --- a type of artificial intelligence system, built as a large neural network trained on broad text data, that generates text in response to a prompt. The two models evaluated in this study are OpenAI's GPT-5 and Google's Gemini 3 Flash. & Bommasani et al. (2021)\tabularnewline
\textbf{Semantic repeatability} & How similar in meaning the LLM's answers are when the exact same question (prompt) is asked multiple times. In this study, each question is asked 30 times; every pair of answers is compared for meaning, and those similarities are averaged. The score runs 0--1, where higher means more consistent. & Shyr et al. (2025)\tabularnewline
\textbf{Semantic reproducibility} & How similar in meaning the LLM's answers are when the same underlying question is asked in 5 different ways (reworded as Chain-of-Thought (CoT) variants). For each CoT variant, we average its 30 LLM responses, defining what we call a centroid for that variant; the 5 centroids for the same question are then combined into a single prompt-reproducibility score by averaging their pairwise meaning-similarities (step 5 in Figure 1). The score runs 0--1, where higher means more consistent across phrasings. & Shyr et al. (2025)\tabularnewline
\textbf{Centroid (per CoT variant)} & The average of all 30 answer-vectors for a given CoT wording. & Shyr et al. (2025)\tabularnewline
\textbf{Chain-of-Thought (CoT) prompting} & A prompting strategy that frames a question to elicit intermediate reasoning steps from the LLM before the final answer; different CoT framings of the same underlying question constitute the 5 variants used in the reproducibility analysis. & Savage et al. (2024)\tabularnewline
\textbf{Sentence embedding} & A list of numbers (vectors) that captures the meaning of a piece of text, produced by a transformer model. Two texts that mean similar concepts are represented by vectors that point in similar directions. Throughout this paper, the terms \emph{``embedding model''}, \emph{``sentence-embedding model''}, and \emph{``transformer''} (or \emph{``sentence-transformer''}) are used interchangeably to refer to the model that produces these embeddings & Reimers \& Gurevych (2019); Song et al. (2020)\tabularnewline
\textbf{Cosine similarity} & A measure of orientation similarity between two vectors, ranging from $-1$ (opposite) to 1 (identical). In this study, cosine similarities are rescaled to {[}0, 1{]} using $\tilde{S} = (\bar{S} + 1) / 2$ for interpretability. & Shyr et al. (2025)\tabularnewline
\textbf{Temperature} & A sampling hyperparameter of the LLM decoder that controls randomness in token selection; higher values produce more variable outputs. In both the OpenAI (GPT-5) and Google (Gemini 3 Flash) APIs, temperature ranges from 0 to 2, with a default of 1.0. This is also the setting applied in most public-facing chat interfaces available. We used temperature = 1.0, so our results reflect how the models behave in typical real-world settings. & OpenAI (2026); Google (2026)\tabularnewline
\bottomrule
\end{longtable}
}\addtocounter{table}{-1}

\subsection*{2.2 Prompt corpus}

Fifteen H5N1-related questions were curated from public discussion on Reddit during the H5N1 outbreak period (step 1 in Figure 1). Questions were selected manually by browsing publicly available community discussions and span a range of One Health topics, including viral transmission, food safety, household pets, agricultural practices, and public-health response. Under Reddit's Public Content Policy (Reddit, 2026), public Reddit content is made available for non-commercial learning and research and does not require prior authorization from Reddit for use of the kind performed here. The study was additionally covered by the Cornell Institutional Review Board notice of exemption (\#IRB0150253, dated Oct 29, 2026).

Aligned with Reddit policies (Reddit, 2026), to protect contributor privacy, (i) no identifying metadata (usernames, user IDs) was collected or retained at any point in this study; (ii) original source URLs, exact post wording, dates, and the names of the specific subreddits browsed are not disclosed in this manuscript; and (iii) the prompt corpus used in the LLM queries is not released.

The 15 base questions reflecting the prompt corpus are listed below (Table 2). All items are paraphrased reformulation\textbf{s} of the original Reddit posts, preserving the semantic content of each original while differing in surface form so that reverse-text search cannot recover the source post. Each question is tagged with its underlying One Health theme (Table 2); the 15 questions collectively span human transmission, companion animals, agricultural policy and surveillance, food safety, carcass disposal, vaccine development, pandemic preparedness, and viral pathogenesis. The question numbering in Table 2 is retained throughout the manuscript so that specific questions can be referenced by index (e.g., Q5) in the Results and Discussion.

\refstepcounter{table}\label{tab:questions}%
\noindent Table 2. The 15 social-media questions prompted to LLMs.

{\small
\begin{longtable}[]{@{}>{\raggedright\arraybackslash}p{0.09\linewidth}>{\raggedright\arraybackslash}p{0.60\linewidth}>{\raggedright\arraybackslash}p{0.25\linewidth}@{}}
\toprule
Question Number & Question & One Health Theme\tabularnewline
\midrule
\endhead
1 & \emph{Why have the number of reported human H5N1 cases declined over time?} & public-health surveillance\tabularnewline
2 & \emph{Is culling entire poultry flocks the most effective strategy for containing H5N1 outbreaks?} & agricultural policy / disease control\tabularnewline
3 & \emph{If H5N1 were to trigger the next pandemic, how long might it be expected to last?} & pandemic preparedness\tabularnewline
4 & \emph{Is there a risk associated with giving freeze-dried snacks to cats during an active H5N1 outbreak?} & companion-animal food safety\tabularnewline
5 & \emph{Can H5N1 be transmitted from mother to infant through breastfeeding?} & maternal/vertical transmission\tabularnewline
6 & \emph{Could cats serve as sentinel species for H5N1 in local communities, and what response actions should follow a positive case?} & veterinary surveillance / companion animals\tabularnewline
7 & \emph{Is thoroughly cooked meat from H5N1-infected poultry safe for human consumption?} & food safety\tabularnewline
8 & \emph{Could the apparent decline in H5N1 cases be explained by reduced testing or surveillance?} & epidemiologic interpretation\tabularnewline
9 & \emph{Why is testing of dairy cattle delayed until clinical symptoms appear when H5N1 is already known to be circulating in herds?} & agricultural surveillance\tabularnewline
10 & \emph{What precautions can cat owners take to reduce their pets' risk during an H5N1 outbreak?} & companion-animal care\tabularnewline
11 & \emph{Why aren't the carcasses of H5N1-infected birds incinerated to prevent further spread of the virus?} & disease control\tabularnewline
12 & \emph{Is H5N1 transmission from one person to another possible?} & human-to-human transmission\tabularnewline
13 & \emph{Could mass culling of chickens in the United States create long-term reliance on imported chicks from other countries?} & agricultural economics\tabularnewline
14 & \emph{Could the annual seasonal influenza vaccine simply be reformulated to protect against H5N1?} & vaccine development\tabularnewline
15 & \emph{Why is H5N1 infection lethal to some species but mild in others?} & viral pathogenesis / host range\tabularnewline
\bottomrule
\end{longtable}
}\addtocounter{table}{-1}

The corpus from Table 2 was used in two complementary semantic consistency analyses, defined in Table 1 and summarized visually in Figure 1 (step 2):

\begin{itemize}
\item
  Semantic repeatability analysis --- each of the 15 questions was submitted 30 times under identical conditions, yielding 450 responses per model.
\item
  Semantic reproducibility analysis --- each of the 15 questions was reformulated into 5 CoT reasoning variants, and each variant was run 30 times (5 $\times$ 30 = 150 responses per question), yielding 2,250 responses per model.
\end{itemize}

The five CoT framings used in the reproducibility analysis follow the five-prompt taxonomy originally developed by Savage et al. (2024) for clinical diagnostic reasoning and subsequently adopted into the semantic consistency framework of Shyr et al. (2025). These variants preserve the underlying question while varying the elicited reasoning style. Table 3 shows the generic template for each CoT reasoning frame alongside a worked example in which base question Q5 (``Can H5N1 be transmitted from mother to infant through breastfeeding?'') is expanded into each of the five CoT variants.

\refstepcounter{table}\label{tab:cot}%
\noindent \textbf{Table 3. Chain-of-Thought (CoT) reasoning frames used in the reproducibility analysis.}

{\small
\begin{longtable}[]{@{}>{\raggedright\arraybackslash}p{0.18\linewidth}>{\raggedright\arraybackslash}p{0.33\linewidth}>{\raggedright\arraybackslash}p{0.43\linewidth}@{}}
\toprule
CoT variant & Reasoning-frame template & Example applied to Q5\tabularnewline
\midrule
\endhead
v1 --- Traditional CoT & \emph{``Think step by step to answer\ldots"} & \emph{``Think step by step to answer whether H5N1 can be transmitted from mother to infant through breastfeeding."}\tabularnewline
v2 --- Differential Diagnosis CoT & \emph{"Compare possible options/explanations/possibilities to determine\ldots"} & \emph{``Compare possible transmission routes to determine whether H5N1 can be transmitted from mother to infant through breastfeeding."}\tabularnewline
v3 --- Intuitive Reasoning CoT & \emph{"Consider historical cases to intuitively infer\ldots"} & \emph{"Consider historical patterns of viral transmission through breast milk to intuitively infer whether H5N1 can be transmitted from mother to infant through breastfeeding."}\tabularnewline
v4 --- Analytic Reasoning CoT & \emph{"Analyze the mechanics behind cases and answer\ldots"} & \emph{"Analyze the mechanics behind viral transmission through breast milk and answer whether H5N1 can be transmitted from mother to infant through breastfeeding."}\tabularnewline
v5 --- Bayesian Reasoning CoT & \emph{"To answer\ldots, start with a prior belief about it, update it with evidence, and provide a conclusion."} & \emph{"To answer whether H5N1 can be transmitted from mother to infant through breastfeeding, start with a prior belief about viral transmission in breast milk, update it with evidence, and provide a conclusion."}\tabularnewline
\bottomrule
\end{longtable}
}\addtocounter{table}{-1}

\subsection*{2.3 LLM response generation}

Two commercially available LLMs were queried via their official APIs:

\begin{itemize}
\item
  \begin{quote}
  GPT-5 (OpenAI; model ID gpt-5; released August 7, 2025)
  \end{quote}
\item
  \begin{quote}
  Gemini 3 Flash Preview (Google; model ID gemini-3-flash-preview; released December 17, 2025)
  \end{quote}
\end{itemize}

These models were selected to represent two major LLM providers' widely used offerings during the H5N1 outbreak period, with similar release dates (approximately four months apart). All API calls used identical configuration parameters across both models to ensure protocol consistency:

\begin{itemize}
\item
  \begin{quote}
  Temperature = 1.0 --- controls output randomness during token sampling; this value was selected to reflect the default consumer-facing experience (explained in Table 1).
  \end{quote}
\item
  \begin{quote}
  Maximum concurrent requests = 5 (asyncio semaphore).
  \end{quote}
\item
  \begin{quote}
  Inter-run spacing = 60 seconds within the same prompt/variant.
  \end{quote}
\end{itemize}

For the repeatability analysis, each of the 15 prompts was issued to each model 30 times under identical conditions, producing 450 responses per model (900 responses total). For the reproducibility analysis, each of the 75 variant prompts (15 prompts $\times$ 5 variants) was issued 30 times per model, producing 2,250 responses per model (4,500 responses total) (see Figure 1 Step 3). The complete corpus comprises 5,400 LLM responses (see Table 4).

\refstepcounter{table}\label{tab:design}%
\noindent \textbf{Table 4. Response collection design.}

{\small
\begin{longtable}[]{@{}>{\raggedright\arraybackslash}p{0.15\linewidth}>{\raggedright\arraybackslash}p{0.09\linewidth}>{\raggedright\arraybackslash}p{0.17\linewidth}>{\raggedright\arraybackslash}p{0.17\linewidth}>{\raggedright\arraybackslash}p{0.14\linewidth}>{\raggedright\arraybackslash}p{0.11\linewidth}@{}}
\toprule
Analysis & Prompts & Variants per prompt & Runs per (prompt, variant) & Responses per model & Total responses\tabularnewline
\midrule
\endhead
Repeatability & 15 & 1 base & 30 & 450 & 900\tabularnewline
Reproducibility & 15 & 5 variants of the base & 30 & 2,250 & 4,500\tabularnewline
Combined corpus & --- & --- & --- & 2,700 & 5,400\tabularnewline
\bottomrule
\end{longtable}
}\addtocounter{table}{-1}

\subsection*{2.4 Semantic consistency framework}

Semantic consistency scores were computed using an adapted version of the statistical framework proposed by \citet{Shyr2025}. We extend their methodology in four ways: (i) we apply the framework to naturalistic public-health questions rather than clinical vignettes; (ii) we substitute a general-purpose sentence-embedding model for their clinical-domain embedding; (iii) we extend the framework to cross-provider benchmarking across two commercial LLMs from different companies (OpenAI and Google), and (iv) we extend the statistical pipeline with paired Wilcoxon tests (within-model repeatability vs. reproducibility), Friedman tests with post-hoc pairwise Wilcoxon and Holm--Bonferroni correction (across CoT variants), bootstrap 95\% confidence intervals on per-prompt means, and ICC(2,k) and Cronbach's $\alpha$ for inter-embedding agreement.

\begin{figure}[H]
\centering
\includegraphics[width=\textwidth]{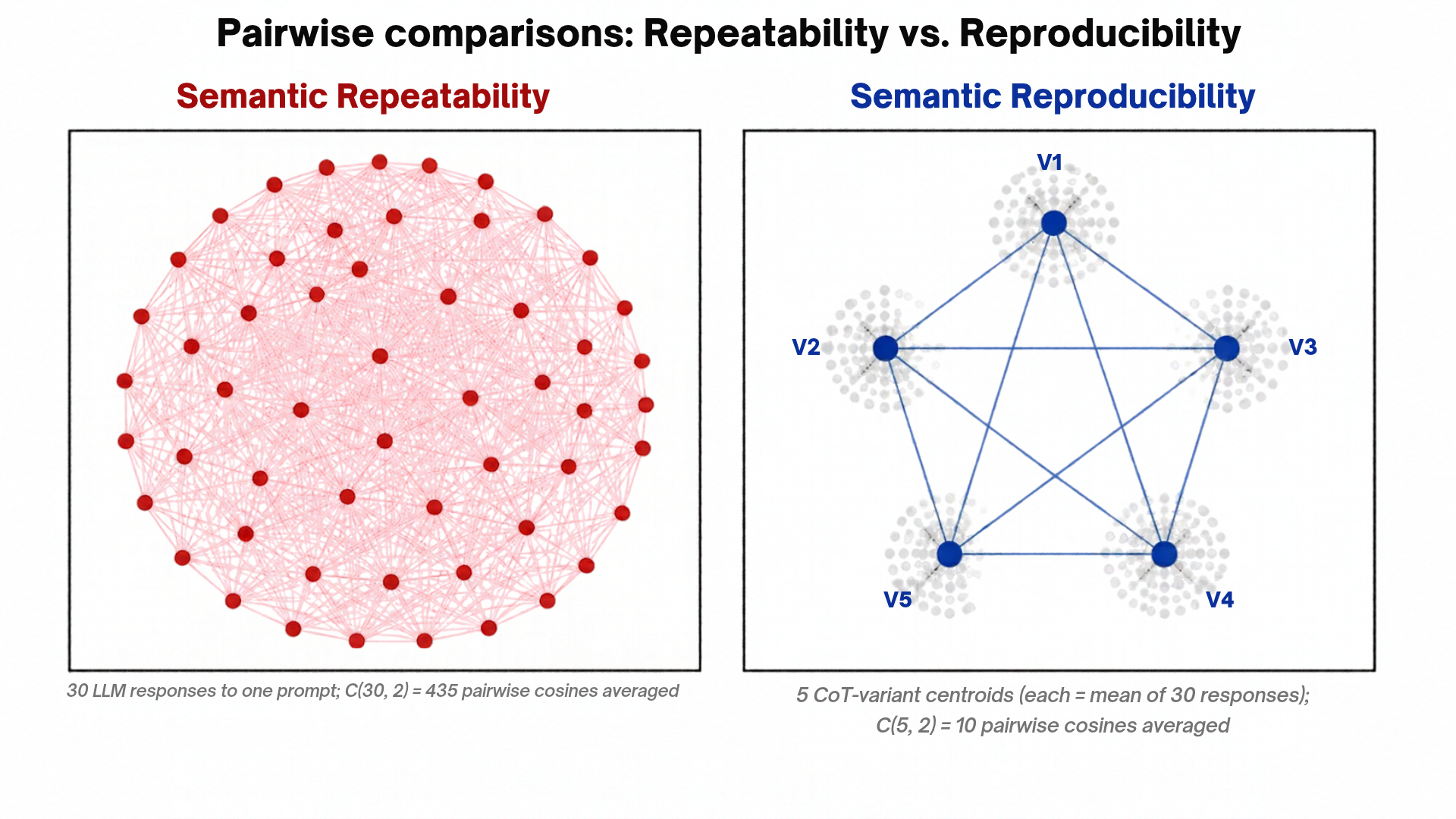}
\caption{\textbf{Figure 2. Pairwise comparison: Repeatability vs.\ Reproducibility.} Structural asymmetry between the two pairwise-similarity calculations. Left: semantic repeatability for a single prompt is the mean of $C(30, 2) = 435$ pairwise cosine similarities among the 30 LLM responses to that prompt. Right: semantic reproducibility for the same prompt is the mean of $C(5, 2) = 10$ pairwise cosine similarities among the 5 CoT-variant centroids, where each centroid is itself the mean of 30 response embeddings (small faint dots in the background).}
\label{fig:pairwise}
\end{figure}

Semantic repeatability (Figure~\ref{fig:pipeline} Step~5) was measured by the mean pairwise cosine similarity among the 30 responses to each identical prompt (Figure~\ref{fig:pairwise}) using the equation:

\begin{equation}
\Sbar_{\Rpt,q} \;=\; \frac{2}{R(R-1)} \sum_{1 \le r < s \le R} \cos(e_r, e_s)
\label{eq:repeatability}
\end{equation}

The symbols in the Equation~\ref{eq:repeatability} are defined as follows:
\begin{itemize}
\item $\Sbar_{\Rpt,q}$ is the semantic repeatability score for question $q$ --- i.e., a single consistency score summarizing how similar the LLM's 30 responses to question $q$ are to one another, on average.
\item $R = 30$ is the number of LLM responses we collected for question $q$.
\item $r$ and $s$ are labels that identify two different responses from that set of 30. The condition $r < s$ in the summation means we count every unique pair of responses exactly once --- e.g., we compare response 3 with response 7, but we do not also compare response 7 with response 3.
\item $e_r$ and $e_s$ are the sentence embeddings of those two responses. A sentence embedding is a numerical vector representation of a text's meaning, produced by a sentence-transformer model.
\item $\cos(e_r, e_s)$ is the cosine similarity between the two embeddings --- a number between $-1$ and $1$ that quantifies how semantically similar the two responses are. Values near 1 indicate near-identical meaning; values near $-1$ indicate opposite meaning; values near 0 indicate unrelated content.
\end{itemize}

Interpretation of the equation output: for each question $q$, we compute the cosine similarity between every unique pair of the 30 LLM responses --- $C(30, 2) = 435$ unique pairs per prompt --- then take the average. That average is the semantic repeatability score for question $q$.

Semantic reproducibility (Figure~\ref{fig:pipeline} Step~5) was measured using a centroid-based approach: for each question $q$, we first averaged the 30 response embeddings of each CoT variant into a single centroid vector (Equation~\ref{eq:centroid}), then computed the cosine similarity between every pair of the 5 centroids and averaged those $C(5, 2) = 10$ similarity measures to produce the per-prompt reproducibility score (Equation~\ref{eq:reproducibility}) (Figure~\ref{fig:pairwise}). This formulation follows \citet{Shyr2025} exactly.

\begin{equation}
\ebar_v \;=\; \frac{1}{R} \sum_{r=1}^{R} e_r^{\,v}, \qquad v = 1, \dots, V
\label{eq:centroid}
\end{equation}

where:
\begin{itemize}
\item $\ebar_v$ is the centroid (mean embedding) for CoT variant of question $q$ (defined in Table~\ref{tab:concepts}).
\item $R = 30$ is the number of LLM responses we collected for each CoT variant of question $q$.
\item $r$ is a label that identifies one of those $R$ responses, with $r = 1, 2, \dots, R$.
\item $e_r^{\,v}$ is the sentence embedding of the $r$-th response generated under CoT variant $v$ (same as $v$).
\item $v$ is a label that identifies one of the five CoT variants, with $v = 1, \dots, V$ (5 in total).
\end{itemize}

\begin{equation}
\Sbar_{\Rpd,q} \;=\; \frac{2}{V(V-1)} \sum_{1 \le v < v' \le V} \cos(\ebar_v, \ebar_{v'})
\label{eq:reproducibility}
\end{equation}

where:
\begin{itemize}
\item $\Sbar_{\Rpd,q}$ is the semantic reproducibility score for question $q$ --- i.e., a single consistency score summarizing how similar the LLM's responses are, on average, when the same underlying question is asked using different Chain-of-Thought reasoning framings.
\item $V = 5$ is the number of CoT variants per question (v1--v5).
\item $v$ and $v'$ identify two different CoT variants from the set of 5. The condition $1 \le v < v' \le V$ in the summation means we count every unique pair of variants exactly once --- e.g., we compare variant 2 with variant 4, but we do not also compare variant 4 with variant 2.
\item $\ebar_v$ and $\ebar_{v'}$ are the centroid embeddings of CoT variants $v$ and $v'$.
\item $\cos(\ebar_v, \ebar_{v'})$ is the cosine similarity between the two centroid vectors.
\end{itemize}

Interpretation of the equation output: for each question, we first collapse each CoT variant's 30 responses into a single centroid embedding by averaging. We then compute the cosine similarity between every pair of centroids (10 pairs total) and take the average of those 10 similarities. That average is the semantic reproducibility score for question $q$.

Both resulting scores were rescaled to the $[0, 1]$ interval using the transformation
\begin{equation}
\tilde{S} \;=\; \frac{\Sbar + 1}{2}
\label{eq:rescale}
\end{equation}
to facilitate interpretation, following \citet{Shyr2025}.

Figure~\ref{fig:pairwise} illustrates the pairwise comparisons that each score (repeatability and reproducibility) aggregates. The repeatability score for a single prompt averages cosine similarities across all $C(30, 2) = 435$ pairs of LLM responses (Equation~\ref{eq:repeatability}), while the reproducibility score for the same prompt averages cosine similarities across only $C(5, 2) = 10$ pairs of CoT-variant centroids (Equation~\ref{eq:reproducibility}). This difference in underlying pair count means that per-prompt repeatability estimates aggregate substantially more information than per-prompt reproducibility estimates and is one reason why single-prompt reproducibility values can be inherently noisier (less precise) than their repeatability counterparts.

\subsection*{2.5 Sensitivity analysis: stopwords and embedding models}

To ensure that our semantic consistency findings are robust rather than artifacts of two upstream methodological choices, we conducted sensitivity analyses along two dimensions of the embedding pipeline (step 4 in Figure 1): (i) the preprocessing applied to response texts prior to embedding (with vs. without stopword removal), and (ii) the specific sentence-transformer model used to generate the embeddings.

(i) Response texts were preprocessed by removing standard English stopwords prior to embedding, and as a sensitivity analysis, the full pipeline was additionally run without the stopword-removal preprocessing step (i.e., with stopwords preserved in the text passed to the embedding model) for both GPT-5 and Gemini 3 Flash across all four embedding models; results of this comparison, which examine whether semantic consistency scores vary with stopword removal, are reported in the Supplementary Materials (Supplementary S1 Figure, Supplementary S1 Table).

(ii) Primary analyses used the all-mpnet-base-v2 sentence-transformer model (Song et al. 2020; 110M parameters), selected as a medium-tier general-purpose embedding model widely used in semantic similarity research (Ortiz Martes et al., 2025). We used three additional embedding models spanning the small/medium/large/commercial tier spectrum:

\begin{itemize}
\item
  \begin{quote}
  all-MiniLM-L6-v2 (22M parameters; Wang et al., 2020)
  \end{quote}
\item
  \begin{quote}
  bge-large-en-v1.5 (335M parameters; Xiao et al., 2024)
  \end{quote}
\item
  \begin{quote}
  text-embedding-3-large (\textasciitilde3B parameters; OpenAI, 2024)
  \end{quote}
\end{itemize}

Inter-transformer agreement was quantified using ICC(2,k) and Cronbach's $\alpha$. ICC values above 0.75 indicate good agreement, and values above 0.90 indicate excellent agreement (Koo \& Li, 2016).

\subsection*{2.6 Statistical analyses}

Five complementary statistical tests (Figure 1 Step 6) were applied to characterize different aspects of consistency:

(1) Paired Wilcoxon signed-rank tests compared per-prompt repeatability versus reproducibility scores within each model (Figure 1 Step 6; Wilcoxon, 1945). This non-parametric test was selected because the sample of 15 prompts does not justify assumptions of normality. For each of the 15 base questions, we had two scores per model --- one repeatability score and one reproducibility score --- giving 15 paired observations per model. For each base question, the test takes the paired difference between the two metrics (repeatability minus reproducibility), ranks those 15 differences by magnitude, and compares the sum of ranks of positive differences against the sum of ranks of negative differences. If the two sums are roughly balanced, there is no systematic difference between the two metrics; if one side dominates, the two metrics differ systematically, with the dominant side indicating which metric tends to be higher across prompts. A significant result therefore flags a systematic difference between a model's repeatability and reproducibility scores across the 15 prompts.

(2) 95\% confidence intervals (CIs) were computed to quantify the uncertainty around reported statistics. In this study, two different CI methods were used depending on the statistic. For the per-question mean repeatability and reproducibility scores, we computed bootstrap 95\% CIs with 10,000 resamples (Efron,1979). In plain terms: for each combination of model (GPT-5, Gemini 3 Flash) and metric (repeatability, reproducibility), we had 15 per-prompt scores. Bootstrapping resamples these 15 scores with replacement --- meaning each draw of 15 can include the same prompt more than once and leave others out, simulating what we might have observed with a different sample of 15 questions --- 10,000 times, recomputing the mean each time. The 95\% CI is then the range spanning the 2.5th to 97.5th percentiles of those 10,000 recomputed means. A separate 95\% CI method applies to the ICC(2,k) values reported later (test (5) below): those CIs are analytic intervals produced directly by the standard ICC calculation (pingouin package) rather than by bootstrap resampling. Both methods target 95\% coverage but are computed differently.

(3) Friedman tests were applied to the per-(prompt, variant) reproducibility scores (15 prompts $\times$ 5 variants, per model) to assess whether semantic reproducibility differed systematically across the five CoT reasoning-style variants (Figure 1 Step 6; Friedman, 1937). The Friedman test is a non-parametric repeated-measures test appropriate for non-normally distributed scores. In plain terms: for each model, we had a 15 $\times$ 5 grid of reproducibility scores (one score per base question $\times$ CoT variant). Within each prompt, the five variants are ranked from lowest to highest score, and the test asks: does one variant keep showing up at the top (or at the bottom) of the rankings across the 15 prompts more often than chance alone would explain? A significant result therefore indicates that at least one of the five CoT variants produces reproducibility scores that systematically differ from the others across the 15 questions.

(4) Post-hoc pairwise Wilcoxon signed-rank tests with Holm--Bonferroni correction were applied whenever the Friedman test (test (3) above) returned a significant result, to identify which specific CoT variant pairs differed (Figure 1 Step 6; Holm, 1979). The Friedman test only indicates that at least one CoT variant differs from the others, not which pairs differ. To localize the effect, for each model we compared all ten unordered CoT-variant pairs (v1--v5) pairwise using paired Wilcoxon signed-rank tests on the per-prompt reproducibility scores, yielding ten raw p-values per model. Running ten tests, however, inflates the chance of a false positive, so we applied the Holm--Bonferroni correction: we sorted the ten p-values from smallest to largest, then multiplied the smallest by 10, the second-smallest by 9, and so on. By design, this correction controls the family-wise (overall) false-positive rate at our chosen $\alpha$ = 0.05, regardless of how many pairs are compared.

\textbf{(5)} ICC(2,k) and Cronbach's $\alpha$ were used to quantify how well the four embedding models (Section 2.5) agree with each other on per-prompt consistency scores (Figure 1 Step 6; Shrout \& Fleiss, 1979; Cronbach, 1951). For each combination of model (GPT-5, Gemini 3 Flash) and metric (repeatability, reproducibility), each embedding model independently encoded the LLM's responses and produced one consistency score per prompt, yielding a 15 $\times$ 4 grid (prompts $\times$ embedding models); both statistics ask whether the four embedding models arrive at similar consistency scores for the same LLM outputs. In practice, both fall between 0 and 1, where higher values indicate more agreement, and ICC(2,k) values \textgreater{} 0.90 are usually considered excellent (Koo \& Li, 2016). ICC(2,k) is defined (Shrout \& Fleiss, 1979) as:

\begin{equation}
\mathrm{ICC}(2,k) \;=\; \frac{\sigma^{2}_{\text{prompt}}}{\sigma^{2}_{\text{prompt}} + \dfrac{\sigma^{2}_{\text{embedding}} + \sigma^{2}_{\text{error}}}{k}}
\label{eq:icc}
\end{equation}

where $\sigma^{2}_{\text{prompt}}$ is the variance in per-prompt consistency scores across the 15 prompts, $\sigma^{2}_{\text{embedding}}$ is the variance across the four embedding models, $\sigma^{2}_{\text{error}}$ is residual variance, and k = 4 is the number of embedding models averaged. In the notation ICC(2,k), the '2' indicates a two-way random-effects model: both the raters (our four embedding models) and the subjects (our 15 prompts) are treated as random samples from larger populations. Reporting both statistics is a methodological addition relative to Shyr et al. (2025), who used only one embedding model, and provides a check on whether our consistency findings depend on the specific embedding model chosen.

All statistical analyses were performed in Python 3.11.3 using: scipy.stats (Wilcoxon, Friedman), numpy.random (bootstrap), pingouin (ICC). Significance threshold was set at $\alpha$ = 0.05.

\section*{3. Results}

\subsection*{3.1 Response corpus overview}

The complete response corpus comprised 5,400 LLM responses --- 2,700 per model, split between 450 repeatability responses (15 base questions $\times$ 30 runs) and 2,250 reproducibility responses (15 base questions $\times$ 5 CoT variants $\times$ 30 runs) --- matching the collection target specified in Methods Section 2.3 (Table 4). All 5,400 API calls completed with a normal end-of-response status returned by both providers' APIs, indicating that each response reached a natural stopping point rather than being cut off by a maximum-output-token cap, a safety filter, or an error. No responses were truncated, dropped, or re-collected. Mean response length was 1,586 completion tokens (median 1,556) for GPT-5 repeatability runs and 2,580 tokens (median 2,600) for GPT-5 reproducibility runs; 827 tokens (median 839) for Gemini 3 Flash repeatability runs and 1,041 tokens (median 1,051) for Gemini reproducibility runs. The longer outputs observed under the reproducibility condition for both models reflect the elaborative nature of CoT prompting. Response-length comparisons between GPT-5 and Gemini 3 Flash should not, however, be interpreted as direct cross-provider comparisons: the two models sit at different tiers within their respective product lines (Gemini 3 Flash is Google's lightweight/low-latency tier, whereas GPT-5 is OpenAI's full flagship model).

\subsection*{3.2 Semantic repeatability}

Both GPT-5 and Gemini 3 Flash exhibited consistently high semantic repeatability on H5N1 One Health questions. Across the 15 base questions (Q1--Q15), each issued 30 times, mean per-prompt repeatability under the primary pipeline (MPNet, stopwords removed) was 0.932 for GPT-5 and 0.968 for Gemini 3 Flash on the rescaled {[}0, 1{]} scale (Table 5), indicating that each model produced closely matching outputs under identical-prompt resampling.

\refstepcounter{table}\label{tab:repeatability}%
\noindent \textbf{Table 5.} Semantic repeatability summary statistics (MPNet, stopwords removed, rescaled to {[}0, 1{]}). Mean values are averages of the per-prompt repeatability scores across the 15 base questions. 95\% confidence intervals (CI) are bootstrap intervals (10,000 resamples). Min and max refer to the lowest- and highest-scoring individual base questions, with the corresponding question number in parentheses.

{\small
\begin{longtable}[]{@{}lllll@{}}
\toprule
Model & Mean & 95\% CI & Min (question) & Max (question)\tabularnewline
\midrule
\endhead
GPT-5 & 0.932 & {[}0.918, 0.944{]} & 0.872 (Q8) & 0.967 (Q5)\tabularnewline
Gemini 3 Flash & 0.968 & {[}0.964, 0.971{]} & 0.955 (Q1) & 0.981 (Q3)\tabularnewline
\bottomrule
\end{longtable}
}\addtocounter{table}{-1}

For GPT-5, the lowest-scoring question was Q8 (epidemiologic interpretation), while the highest was Q5 (maternal/vertical transmission). For Gemini 3 Flash, Q1 (declining human-case counts) was the lowest-scoring question, and Q3 (hypothetical pandemic duration) the highest (see Figure 3).

\begin{figure}[H]
\centering
\includegraphics[width=\textwidth]{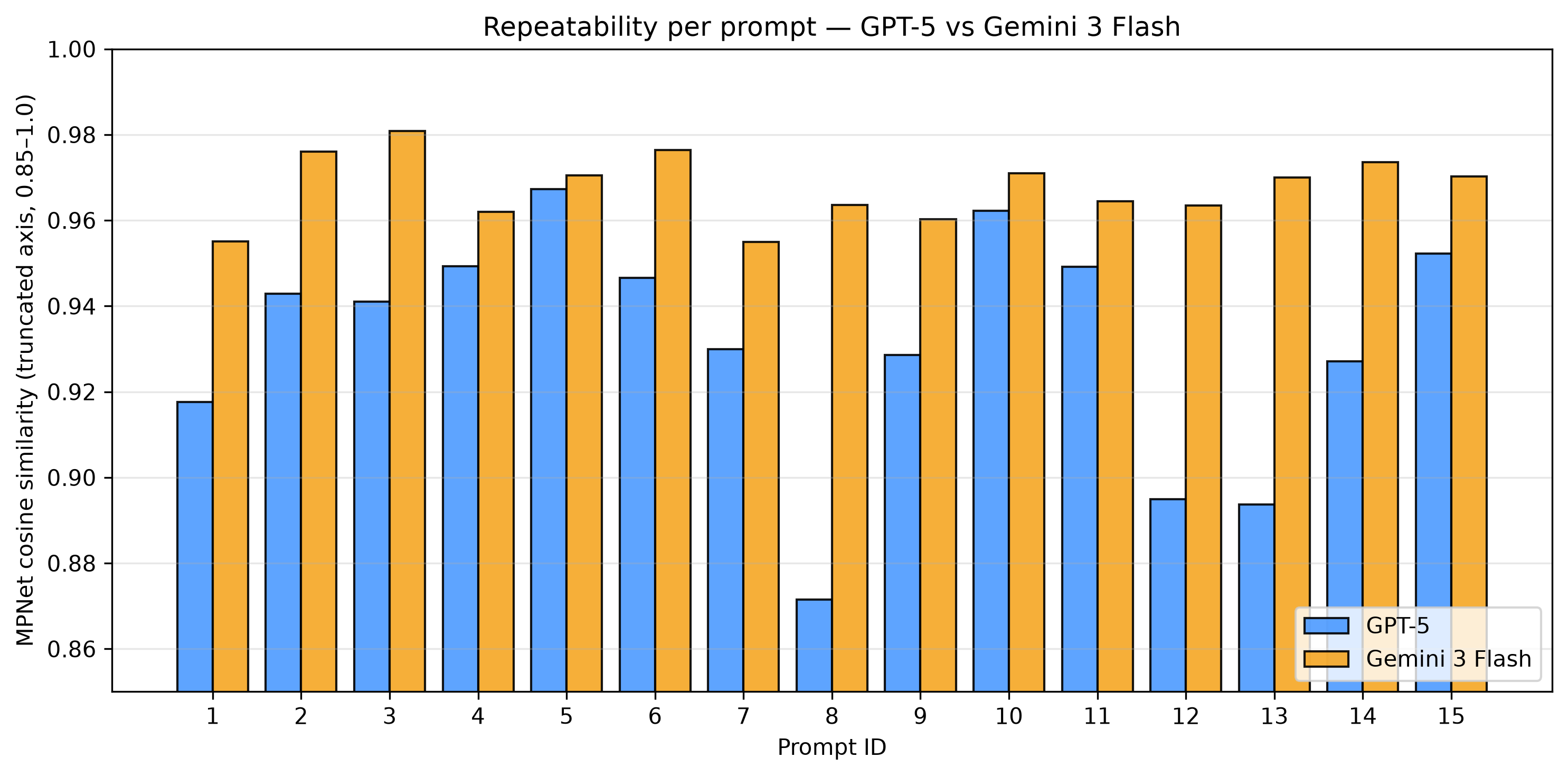}
\caption{\textbf{Figure 3.} Per-prompt mean repeatability scores for GPT-5 and Gemini 3 Flash (MPNet embeddings, stopwords removed). Each bar represents one of the 15 base questions (Q1--Q15). The y-axis is truncated to the 0.85 to 1 range to aid visual interpretation. The within-prompt distribution underlying each bar is shown in Supplementary S4 Figure.}
\label{fig:repeatability}
\end{figure}

\subsection*{3.3 Semantic reproducibility}

Reproducibility --- consistency across CoT reformulations of the same underlying question --- was similarly high for both models, with mean per-prompt reproducibility of 0.962 for GPT-5 and 0.966 for Gemini 3 Flash on the rescaled {[}0, 1{]} scale (Table 6).

\refstepcounter{table}\label{tab:reproducibility}%
\noindent \textbf{Table 6.} Semantic reproducibility summary statistics by model (MPNet, stopwords removed, rescaled to {[}0, 1{]}). Mean values are averages of the per-prompt reproducibility scores across the 15 base questions. 95\% confidence intervals (CI) are bootstrap intervals (10,000 resamples). Min and max refer to the lowest- and highest-scoring individual base questions, with the corresponding question number in parentheses.

{\small
\begin{longtable}[]{@{}lllll@{}}
\toprule
Model & Mean & 95\% CI & Min (question) & Max (question)\tabularnewline
\midrule
\endhead
GPT-5 & 0.962 & {[}0.955, 0.967{]} & 0.934 (Q12) & 0.976 (Q15)\tabularnewline
Gemini 3 Flash & 0.966 & {[}0.961, 0.970{]} & 0.950 (Q12) & 0.986 (Q11)\tabularnewline
\bottomrule
\end{longtable}
}\addtocounter{table}{-1}

For GPT-5, Q12 (``Is H5N1 transmission from one person to another possible?'') was the lowest-scoring question and Q15 (why lethal to some species but mild to others) the highest. For Gemini 3 Flash, Q12 was also the lowest and Q11 (carcass disposal / incineration policy) the highest (see Figure 4).

\begin{figure}[H]
\centering
\includegraphics[width=\textwidth]{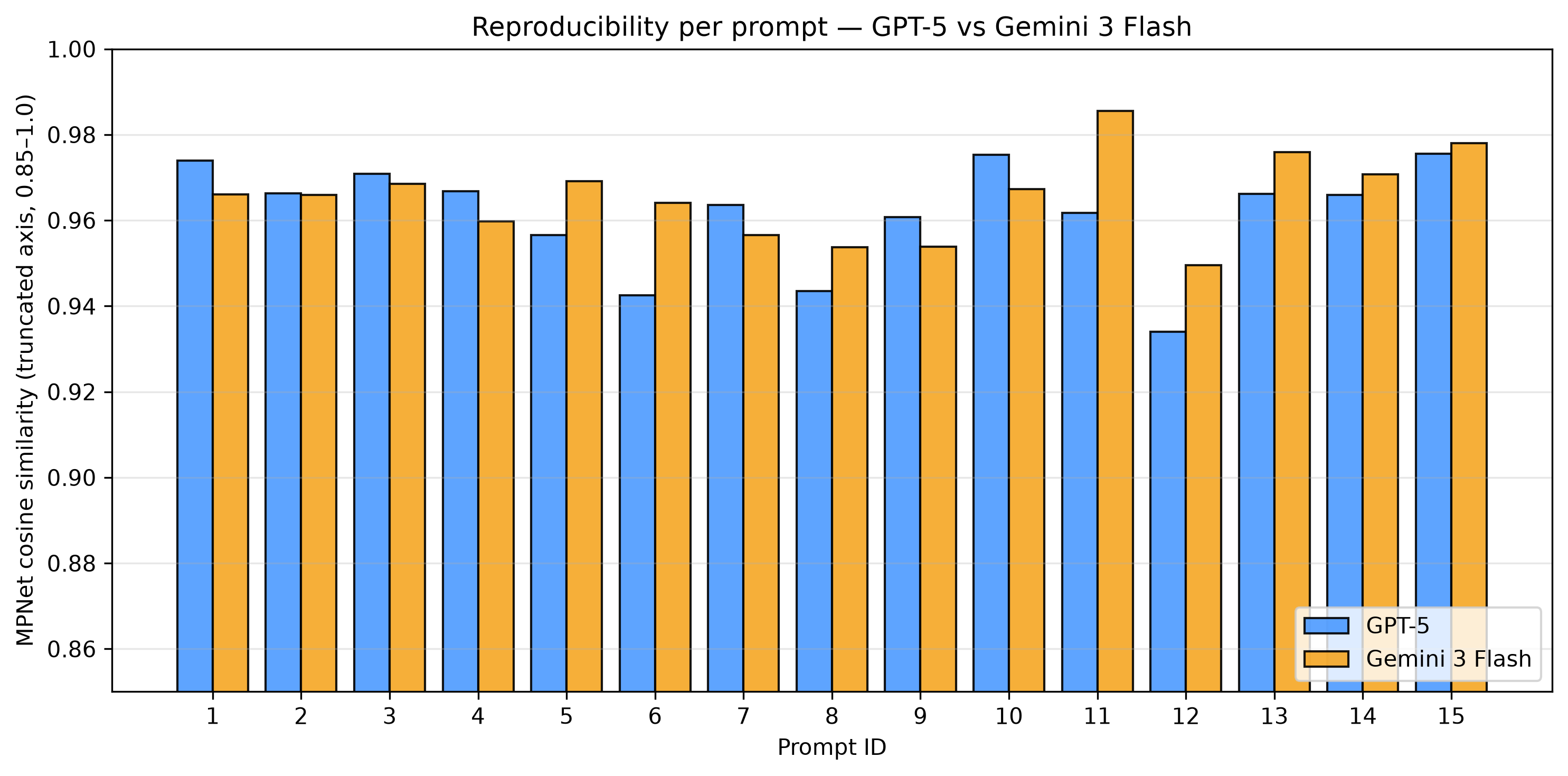}
\caption{\textbf{Figure 4.} Per-prompt mean reproducibility scores for GPT-5 and Gemini 3 Flash (MPNet embeddings, stopwords removed). The y-axis is truncated to the 0.85 to 1 range to aid visual interpretation. The within-prompt distribution among the 5 CoT-variant centroids underlying each bar is shown in Supplementary S5 Figure.}
\label{fig:reproducibility}
\end{figure}

\subsection*{3.4 Repeatability versus reproducibility comparison}

The relationship between repeatability (consistency under identical-prompt resampling) and reproducibility (consistency across CoT reformulations of the same question) differed by model. For GPT-5, reproducibility exceeded repeatability in 13 of 15 base questions, with a mean gap of 0.030 on the rescaled scale (see Table 7 and Figure 5). To confirm that the direction of effect was not an artifact of the specific embedding model or preprocessing choice, we additionally examined the paired comparison across the three supplementary embedding models and the alternative preprocessing pipeline (without stopword removal) specified in Methods Section 2.5 as pre-planned robustness checks. The direction of effect --- reproducibility greater than repeatability --- was confirmed across the majority of configurations, with every stopword-removal configuration reaching significance. GPT-5's per-prompt repeatability scores were also more variable across the 15 prompts than its reproducibility scores (see Figure 5 and Supplementary S6 Figure for the per-prompt distributions; full configuration-level results are reported in Supplementary S2 Table).

For Gemini 3 Flash, repeatability and reproducibility were statistically indistinguishable under the primary pipeline (see Table 7). Sentivity checks across the three supplementary embedding models were consistent with a near-zero underlying effect; full configuration-level results are reported in Supplementary S2 Table.

\refstepcounter{table}\label{tab:wilcoxon}%
\noindent \textbf{Table 7.} Paired Wilcoxon signed-rank test comparing per-prompt repeatability and reproducibility, by model, under the primary analytic pipeline (MPNet, stopwords removed). All means are rescaled to {[}0, 1{]}. Results from the three additional embedding models and the alternative preprocessing pipeline (without stopword removal), tested per model as pre-planned robustness checks, are reported in Supplementary S2 Table.

{\small
\begin{longtable}[]{@{}lllll@{}}
\toprule
Model & Mean Repeatability & Mean Reproducibility & W* & p\tabularnewline
\midrule
\endhead
GPT-5 & 0.932 & 0.962 & 3.0 & 0.0003\tabularnewline
Gemini 3 Flash & 0.968 & 0.966 & 42.0 & 0.33\tabularnewline
\bottomrule
\end{longtable}
}\addtocounter{table}{-1}

*W is the Wilcoxon signed-rank test statistic, which for our n = 15 paired observations takes values between 0 and 60. A small W (close to 0) means almost all 15 prompts favour the same direction, indicating a strong systematic effect; a large W (close to 60) means the differences across prompts are balanced and no systematic direction is detectable.

\begin{figure}[H]
\centering
\includegraphics[width=\textwidth]{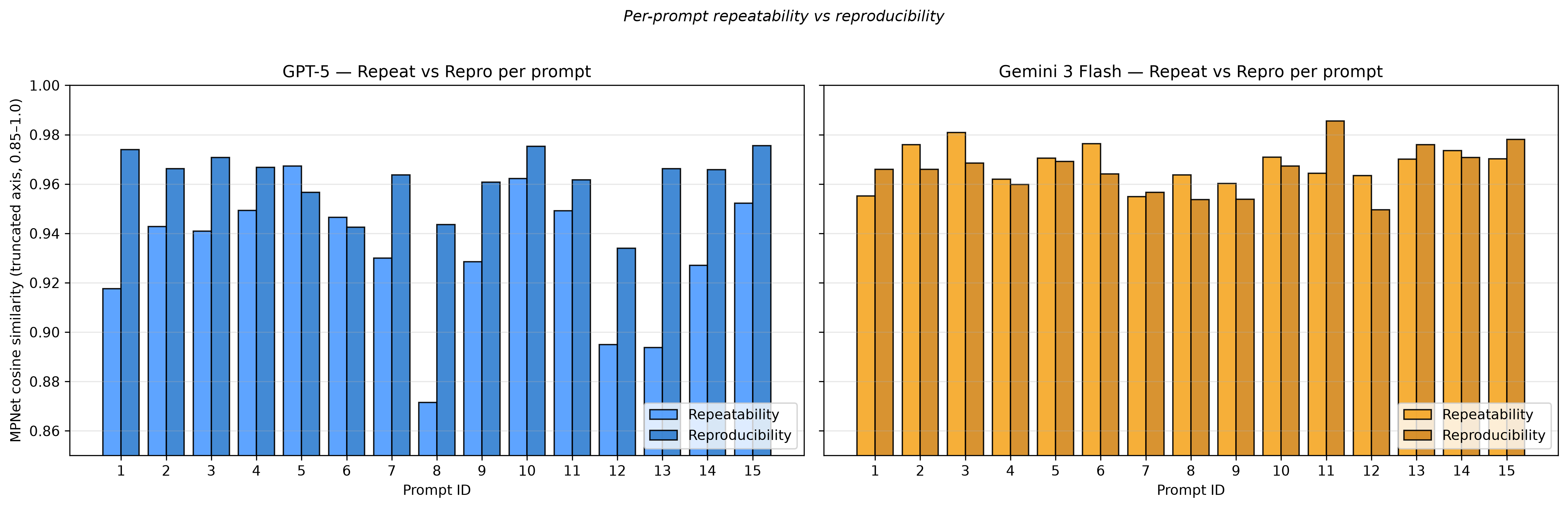}
\caption{\textbf{Figure 5.} Side-by-side comparison of per-prompt repeatability versus reproducibility for each model (MPNet, stopwords removed). The y-axis is truncated to the 0.85 to 1 range to aid visual interpretation. The within-prompt distributions are shown side by side for both models in Supplementary S6 Figure.}
\label{fig:repeat_vs_repro}
\end{figure}

\subsection*{3.5 Variability across Chain-of-Thought reasoning variants}

Not all CoT framings produced equally consistent responses. Within each model, we compared the mean reproducibility score across the five CoT variants (v1--v5). To test whether the choice of CoT framing had a systematic effect on reproducibility --- rather than all five variants producing broadly similar scores --- we used a Friedman test. The test was significant for both models (GPT-5: $\chi^{2}$(4) = 19.04, p = 0.0008; Gemini 3 Flash: $\chi^{2}$(4) = 15.57, p = 0.0036), indicating that at least one of the five CoT variants produces reproducibility scores that differ systematically from the others. Larger $\chi^{2}$ values indicate stronger evidence of systematic between-variant differences; for our design, the critical value at df = 4 and $\alpha$ = 0.05 is approximately 9.49, which both observed statistics exceed. Mean per-variant reproducibility scores for each model are shown in Table 8.

\refstepcounter{table}\label{tab:variants}%
\noindent \textbf{Table 8.} Mean per-prompt reproducibility by Chain-of-Thought variant, for GPT-5 and Gemini 3 Flash (MPNet, stopwords removed, rescaled to {[}0, 1{]}). Each mean is averaged across the 15 base questions. Bold values indicate the highest-scoring variant for each model; italic values indicate the lowest-scoring variant for each model.

{\small
\begin{longtable}[]{@{}>{\raggedright\arraybackslash}p{0.30\linewidth}>{\raggedright\arraybackslash}p{0.11\linewidth}>{\raggedright\arraybackslash}p{0.17\linewidth}>{\raggedright\arraybackslash}p{0.14\linewidth}>{\raggedright\arraybackslash}p{0.17\linewidth}@{}}
\toprule
CoT variant & GPT-5 mean & GPT-5 IQR & Gemini 3 Flash mean & Gemini 3 Flash IQR\tabularnewline
\midrule
\endhead
v1 --- Traditional CoT & \emph{0.9546} & {[}0.9464, 0.9641{]} & 0.9686 & {[}0.9648, 0.9715{]}\tabularnewline
v2 --- Differential Diagnosis CoT & 0.9642 & {[}0.9597, 0.9757{]} & \textbf{0.9702} & {[}0.9665, 0.9773{]}\tabularnewline
v3 --- Intuitive Reasoning CoT & \textbf{0.9657} & {[}0.9599, 0.9743{]} & \emph{0.9609} & {[}0.9539, 0.9699{]}\tabularnewline
v4 --- Analytic Reasoning CoT & 0.9573 & {[}0.9482, 0.9733{]} & 0.9644 & {[}0.9578, 0.9729{]}\tabularnewline
v5 --- Bayesian Reasoning CoT & 0.9620 & {[}0.9588, 0.9707{]} & 0.9649 & {[}0.9614, 0.9683{]}\tabularnewline
\bottomrule
\end{longtable}
}\addtocounter{table}{-1}

All five CoT variants produced consistently high reproducibility scores in both models (all \textgreater{} 0.95; Table 8), indicating strong semantic agreement across reasoning framings. To localize the systematic differences flagged by the Friedman test, we ran paired Wilcoxon signed-rank tests on all 10 variant pairs per model, with Holm--Bonferroni correction applied across the 10 comparisons. In GPT-5, Differential Diagnosis CoT (v2), Intuitive Reasoning CoT (v3), and Bayesian Reasoning CoT (v5) were each significantly more consistent than Traditional CoT (v1) (all Holm-adjusted p $\leq$ 0.034). In Gemini 3 Flash, Differential Diagnosis CoT (v2) was significantly more consistent than Intuitive Reasoning CoT (v3) (Holm-adjusted p = 0.043). Across both models, the Differential Diagnosis CoT (v2) was therefore never significantly outperformed by any other variant. Full post-hoc results for all 20 pairwise comparisons are reported in Supplementary S3 Table.

\begin{figure}[H]
\centering
\includegraphics[width=\textwidth]{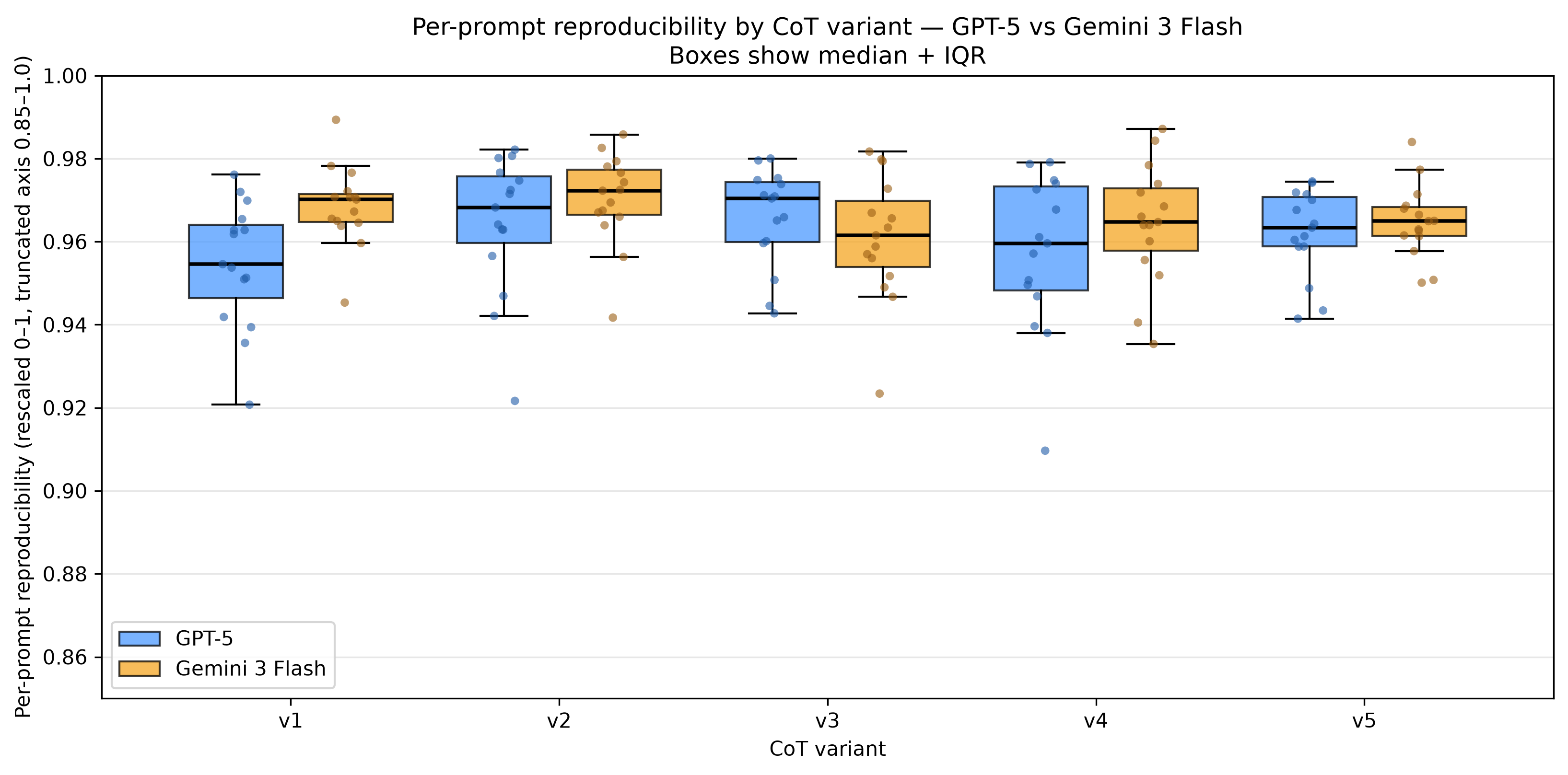}
\caption{\textbf{Figure 6.} Mean per-variant reproducibility (v1--v5) for GPT-5 and Gemini 3 Flash. The y-axis is truncated to the 0.85 to 1 range to aid visual interpretation.}
\label{fig:variant_boxplot}
\end{figure}

All five CoT variants produced tightly clustered per-prompt reproducibility scores in both models, with means between 0.955 and 0.972 and most boxes spanning less than 0.02 units (Figure 6). The variants flagged as significantly less consistent by the post-hoc Wilcoxon tests (v1 Traditional CoT for GPT-5; v3 Intuitive Reasoning CoT for Gemini 3 Flash) stand out in the figure as both the lowest-median boxes and the widest-spread distributions for their respective models. In contrast, Differential Diagnosis CoT (v2) sits at the top of the distribution for both models, consistent with being the only variant that was never significantly outperformed in either model.

\subsection*{3.6 Sensitivity across embedding models}

To confirm that the above findings were not artifacts of the specific sentence-embedding model used, we repeated all per-prompt analyses with three additional embedding models (MiniLM-L6-v2, BGE-large-en-v1.5, OpenAI text-embedding-3-large) and computed ICC(2,k) and Cronbach's $\alpha$ across the four embedding models (see Methods Section 2.5). Inter-transformer agreement was excellent for GPT-5 on repeatability and good on reproducibility, and moderate for Gemini 3 Flash on repeatability and good on reproducibility (Table 9). Per-prompt consistency scores from all four embedding models are plotted side by side in S2 Figure (GPT-5) and S3 Figure (Gemini 3 Flash), providing the visual counterpart to the ICC(2,k) and Cronbach's $\alpha$ summary values reported in Table 9.

\refstepcounter{table}\label{tab:icc}%
\noindent Table 9. Inter-transformer agreement across the four sentence-embedding models (MiniLM-L6-v2, MPNet-base-v2, BGE-large-en-v1.5, OpenAI text-embedding-3-large), computed on the per-prompt consistency scores (stopwords removed, rescaled to {[}0, 1{]}). Intraclass Correlation Coefficient (2,k) interpretation thresholds follow Koo \& Li (2016). The 95\% confidence intervals (CI) shown for ICC(2,k) are formula-based intervals produced by the standard ICC calculation. Full four-transformer per-prompt comparisons and the parallel stopword-removal sensitivity analysis are provided in the Supplementary Materials (S4 Table and S1 Figure).

{\small
\begin{longtable}[]{@{}>{\raggedright\arraybackslash}p{0.14\linewidth}>{\raggedright\arraybackslash}p{0.15\linewidth}>{\raggedright\arraybackslash}p{0.17\linewidth}>{\raggedright\arraybackslash}p{0.15\linewidth}>{\raggedright\arraybackslash}p{0.10\linewidth}>{\raggedright\arraybackslash}p{0.15\linewidth}@{}}
\toprule
Model & Metric & Intraclass Correlation Coefficient (2,k) & 95\% CI & Cronbach $\alpha$ & Agreement Interpretation\tabularnewline
\midrule
\endhead
GPT-5 & Repeatability & 0.940 & {[}0.860, 0.978{]} & 0.953 & Excellent\tabularnewline
GPT-5 & Reproducibility & 0.895 & {[}0.756, 0.961{]} & 0.919 & Good\tabularnewline
Gemini 3 Flash & Repeatability & 0.714 & {[}0.397, 0.890{]} & 0.738 & Moderate\tabularnewline
Gemini 3 Flash & Reproducibility & 0.876 & {[}0.697, 0.955{]} & 0.914 & Good\tabularnewline
\bottomrule
\end{longtable}
}\addtocounter{table}{-1}

The observed difference in inter-transformer agreements on repeatability for GPT-5 (excellent) and Gemini 3 Flash (moderate) was not investigated further in this study, which focuses on characterizing semantic consistency under a primary embedding pipeline rather than on comparing how different embedding models behave on different LLMs. We acknowledge the difference as a methodological caveat: Gemini 3 Flash's per-prompt repeatability scores under the primary MPNet pipeline are more sensitive to the choice of embedding model than GPT-5's, and their absolute values should therefore be interpreted with somewhat more caution. Even so, Gemini 3 Flash's lower repeatability ICC(2,k) (0.714) still sits at the upper end of the `Moderate' range and just below the `Good' threshold of 0.75 (Koo \& Li, 2016), indicating substantive inter-transformer agreement even in this least-agreeing case. All other ICC(2,k) values in Table 9 reach the `Good' or `Excellent' range, supporting the overall robustness of the study's consistency measures.

\section*{4. Discussion}

This study characterized the semantic repeatability and reproducibility of two commercial large language models --- OpenAI's GPT-5 and Google's Gemini 3 Flash --- on 15 naturalistic H5N1 One Health questions sourced from public discourse, using an extension of the statistical framework of Shyr et al. (2025). We report three main findings. First, both models produced consistently high mean semantic consistency scores under both repeated identical prompting and CoT reformulations. Second, the within-model repeatability-vs-reproducibility gap differed systematically between the two providers, and specific CoT framings affected consistency asymmetrically across models. Third, the Shyr et al. framework transferred successfully to a public-health domain, providing, to our knowledge, the first independent empirical demonstration of its portability.

Across the four model-by-metric combinations tested under the primary pipeline, mean per-prompt semantic consistency scores clustered tightly between 0.932 and 0.968 on the rescaled {[}0, 1{]} scale, with the lowest observed per-prompt score reaching 0.87 (GPT-5 repeatability on Q8). Both GPT-5 and Gemini 3 Flash produced semantically consistent outputs on H5N1 One Health questions, with consistency preserved both within the 30 repeated runs of each prompt and across the five reformulations of its underlying question. This finding aligns with prior work showing high semantic consistency across earlier generations of commercial LLMs on different content: Shyr et al. (2025), whose framework we adapted, reported broadly high semantic repeatability and reproducibility scores for ChatGPT-4, ChatGPT-4o-mini, and Llama 3.2-1B on clinical diagnostic reasoning tasks. Hackl et al. (2023), in a different application domain, documented similarly high consistency for GPT-4 when grading the same student responses across multiple time points. To the best of our knowledge, this is the first study to document high semantic consistency for GPT-5 and Gemini 3 Flash specifically on naturalistic H5N1 One Health questions sourced from online discourse. The practical implication is that two current-generation commercial LLMs can be considered semantically stable information sources on H5N1-related questions: a user asking the same question repeatedly, or asking the same underlying question in different ways, is likely to receive answers that preserve the same central meaning. This matters because members of the public increasingly consult LLMs for information about ongoing disease outbreaks (Bautista et al., 2025). An LLM that gave very different answers to the same question would undermine its usefulness as a public-health information touchpoint, and the ability to judge its usefulness and accuracy from a public health perspective (Shyr et al., 2025). Beyond the H5N1 context specifically, our numeric values (mean per-prompt scores of 0.93--0.97 under the primary MPNet pipeline) establish a benchmark range against which future LLM releases and future One Health evaluations can be compared, extending the evaluation reference points published by Shyr et al. (2025) into new health application domains.

Beneath this macro-level consistency, both models showed systematic variability along two axes: the within-model relationship between repeatability and reproducibility, and the ordering of scores across the five CoT reasoning variants. For GPT-5, reproducibility scores systematically exceeded repeatability across 13 of 15 base questions; for Gemini 3 Flash, the two metrics were both high and statistically indistinguishable. This cross-provider divergence aligns with broader prior observations that LLM consistency varies with model, prompt, and dataset. Shyr et al. (2025) reported substantial variation along all three of those axes within their clinical-diagnosis framework. Savage et al. (2024) --- whose five CoT reasoning prompts we adopted and adapted --- found that different reasoning framings produced systematically different diagnostic accuracies on open-ended clinical questions. Li et al. (2024) showed that prompt-engineering style alone can span the full range of inter-rater agreement on orthopedic clinical guidelines --- from essentially no agreement among raters to almost-perfect agreement. Our contribution extends these observations in two ways. First, we provide direct side-by-side documentation that the direction and magnitude of the within-model repeatability-vs-reproducibility gap can differ between two leading commercial LLMs from different providers (OpenAI's GPT-5 vs. Google's Gemini 3 Flash); whereas Savage et al. (2024) and Li et al. (2024) each tested a single provider, our paired design allows this cross-provider contrast itself to be observed. Within the 15 H5N1 One Health questions and the five specific CoT wordings we tested, Differential Diagnosis CoT (v2) was the only reasoning framing to rank among the top two for both models; whether this advantage generalizes to other One Health topics, or to other LLM families, remains to be tested.

Practically, consistency findings reported for one commercial LLM may not transfer uncritically to another, so applied evaluations of LLM consistency may benefit from including more than one provider. Additionally, a prompt framing that performs well on one model does not necessarily perform equally on another: although absolute differences between CoT variants were small (mean per-variant reproducibility 0.955--0.970 across both models; Table 8), the pattern of significant differences diverged by model. In our sample, Traditional step-by-step reasoning (v1) was significantly less consistent only on GPT-5, and Intuitive Reasoning (v3) only on Gemini 3 Flash. These specific framing-by-model interactions could benefit from re-testing across other topic domains, question wordings, and future model updates.

Our third main finding concerns the portability of the analytic framework itself. Shyr et al. (2025) developed their framework for evaluating LLM semantic repeatability and reproducibility on clinical diagnostic reasoning tasks, designing it to work with any suitable embedding model and to be applicable across different content domains, with their own empirical demonstration run on clinical cases using the MedEmbed-Large-v1 embedding model. Our study contributes an empirical demonstration of the same framework applied to a One Health content domain (H5N1 questions sourced from public discourse), evaluated across four general-purpose and commercial sentence-embedding models. The framework's generalizability therefore now has at least one empirical demonstration outside its original clinical application. We also introduce several methodological additions to the Shyr et al. (2025) pipeline (documented in Methods \S{}2.6) that complement the Friedman and post-hoc Wilcoxon tests they used and are designed to be reusable by future adopters. Two other recent LLM-evaluation benchmarks in adjacent spaces --- Fast et al.'s (2024) AMEGA benchmark and Zhou et al.'s (2026) PH-LLM benchmark --- introduced valuable new evaluation infrastructure but are domain-locked and do not provide a general reproducibility framework. Their content is purpose-built for narrow use cases (clinical guideline adherence for AMEGA, public-health infoveillance for PH-LLM), and each benchmark measures task-specific accuracy against its predefined items rather than the semantic consistency of model outputs. Our framework-portability result is therefore complementary to, rather than overlapping with, those benchmarks. More broadly, this makes the framework available to any researcher studying LLM semantic consistency in any content domain, from One health (e.g., mpox, dengue, antimicrobial resistance) to applications beyond it, using the exact pipeline, preprocessing choices, and statistical additions we document here.

Three limitations qualify the interpretation of our findings. First, semantic consistency is not the same as factual correctness: our findings support GPT-5 and Gemini 3 Flash as consistent information sources on H5N1-related questions, but they do not validate the factual accuracy of the underlying claims those models make. A user receiving the same answer every time is not thereby necessarily receiving a correct answer, and consistency testing should be paired with accuracy or expert-review testing before any LLM is recommended as a safe source of public health information. Second, the centroid-based reproducibility calculation we adopt from Shyr et al. (2025) may produce systematically higher scores than methods that preserve within-variant variability. Because each CoT variant's 30 responses are averaged into a single centroid before cross-variant comparison, within-variant sampling noise is smoothed out of the input to the final score; the resulting reproducibility values are therefore specific to this centroid-based formulation and should not be compared directly to reproducibility scores from pairwise or other methods. Third, although we selected the two commercial LLMs based on matched release dates during the H5N1 outbreak period to support cross-provider observation, the two models differ in product tier. GPT-5 is OpenAI's full-tier flagship model, whereas Gemini 3 Flash is Google's lightweight, low-latency variant. Any between-model differences we report --- including the significant within-model repeatability-vs-reproducibility gap observed for GPT-5 but not for Gemini 3 Flash --- cannot be attributed exclusively to provider identity; model tier may also contribute, and isolating provider-level effects from tier-level effects would require a controlled comparison at matched tier (e.g., GPT-5 against Gemini 3 Pro).

Our findings point to four natural extensions. First, longitudinal consistency monitoring would track whether these semantic consistency patterns hold across different provider updates --- commercial LLMs are iterated frequently, and different model versions { } may not behave as consistently as the ones utilized within this study. Second, the same framework could be extended to other emerging zoonotic or pandemic-preparedness topics (e.g., mpox, Nipah virus, antimicrobial resistance) to characterize LLM consistency across the broader One Health space. Third, as noted in the limitations, a matched-tier cross-provider comparison (e.g., GPT-5 against Gemini 3 Pro, or GPT-5 Mini against Gemini 3 Flash) would help isolate provider-level effects from model-tier effects in the cross-provider observations we report here. Fourth, now that semantic consistency has been empirically established for both models on H5N1-related questions, expert-review assessment of the factual accuracy of their responses becomes a natural next step.

\section*{Data and code availability}

Per-prompt consistency scores by model, metric, and embedding model under the primary pipeline are provided as Supplementary S4 Table for full data transparency. The analysis code implementing the semantic consistency framework, including all preprocessing, embedding, statistical, and figure-generation pipelines, is archived on Zenodo (DOI: \href{https://doi.org/10.5281/zenodo.24374226}{10.5281/zenodo.24374226}) and mirrored on GitHub at \url{https://github.com/IvanekLab/GPT5-and-Gemini-3-Flash-Consistency}, and uses Python 3.11.3. Statistical summary outputs (per-prompt consistency scores, bootstrap confidence intervals, ICC(2,k)/Cronbach reliability estimates) are included as Supplementary Tables.

The raw LLM responses are not publicly released due to the Terms of Service of the respective API providers (OpenAI and Google), which restrict redistribution of model outputs. The underlying prompt corpus (15 base questions and 75 Chain-of-Thought variants) is also not released, to protect the privacy of Reddit community contributors. Qualified researchers may contact the corresponding author to request access to either dataset under a data-use agreement that specifies responsible use.

\end{document}